\documentclass{article}
\usepackage{amssymb}
\usepackage{amsmath}
\usepackage{amssymb}
\usepackage[preprint]{corl_2025} 
\usepackage{graphicx} 
\usepackage{amsfonts}
\usepackage{url}
\usepackage{xcolor} 
\usepackage{booktabs}
\usepackage{multirow} 
\usepackage{wrapfig}
\usepackage{caption}  
\usepackage{array}
\usepackage[export]{adjustbox}
\usepackage{siunitx} 
\usepackage{rotating} 
\usepackage{siunitx}  
\usepackage{graphicx}    
\usepackage{subcaption}  

\newcommand{\poor}{\texttimes}
\definecolor{sgreendark}{HTML}{1B9E77} 
\definecolor{sbluedark}{HTML}{7570B3}  
\definecolor{syellowdark}{HTML}{E7298A} 

\title{Omni-Perception: Omnidirectional Collision Avoidance for Legged Locomotion in Dynamic Environments}

\author{
  Zifan Wang$^{1}$, Teli Ma$^{1}$, Yufei Jia$^{4}$, Xun Yang$^{1}$, Jiaming Zhou$^{1}$, 
  \\ \textbf{Wenlong Ouyang$^{1}$, Qiang Zhang$^{1,5}$, Junwei Liang$^{\dag1,2,3}$}
  \\ $^{1}$The Hong Kong University of Science and Technology (Guangzhou)
  \\ $^{2}$Spatialtemporal AI  $^{3}$The Hong Kong University of Science and Technology
  \\ $^{4}$Department of Eletronic Engineering, Tsinghua University 
  \\ $^{5}$Beijing Innovation Center of Humanoid Robotics Co., Ltd.
  \\ {\texttt~\url{https://github.com/aCodeDog/OmniPerception}}
}

\begin{document}
\maketitle
\vspace{-30pt}
\begin{figure*}[h!]
\centering
\includegraphics[width=1.0\linewidth,trim={0cm 0cm 0cm 0cm}]{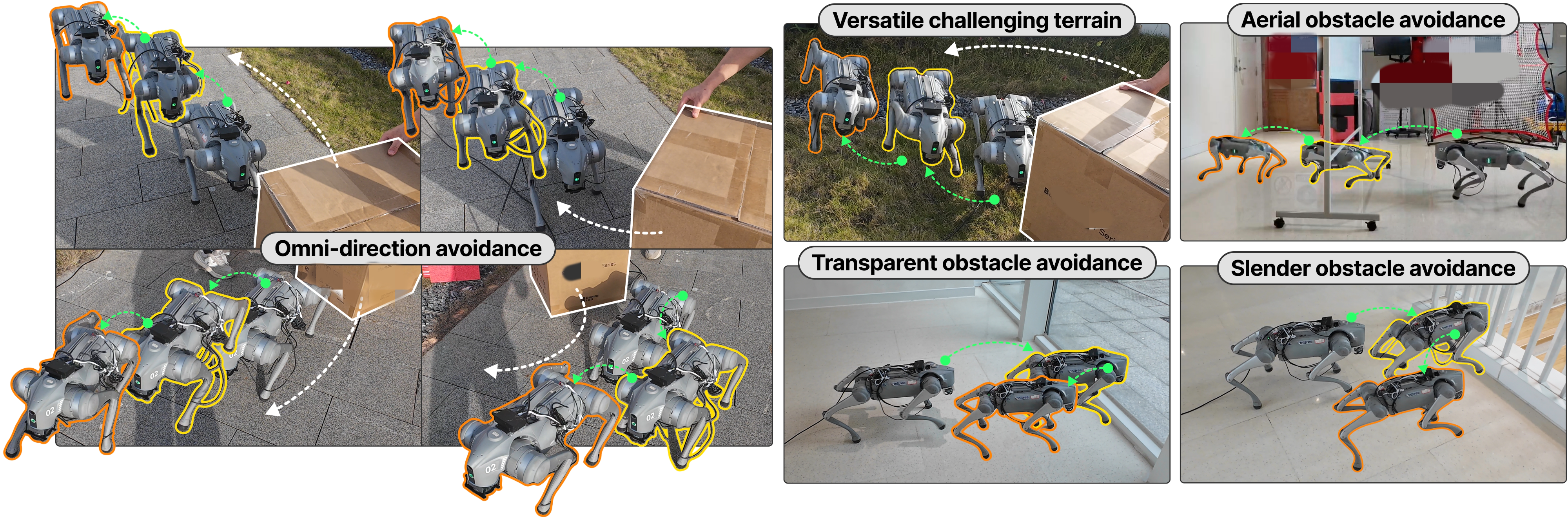}

\caption{\small Validation scenarios for the Omni-Perception framework. Effective omnidirectional collision avoidance is demonstrated on the left, where the robot reacts to obstacles from various approach vectors. Robustness against diverse environmental features is shown on the right, including successful negotiation of aerial, transparent, slender, and ground obstacles. These results highlight the capacity of the Omni-Perceptio to achieve collision-free locomotion in challenging 3D settings directly from raw LiDAR input.} 
\vspace{-10pt}
\label{fig:firstPage}
\end{figure*}

\begin{abstract}
Agile locomotion in complex 3D environments requires robust spatial awareness to safely avoid diverse obstacles such as aerial clutter, uneven terrain, and dynamic agents. Depth-based perception approaches often struggle with sensor noise, lighting variability, computational overhead from intermediate representations. 
In contrast, direct integration of LiDAR sensing into end-to-end learning for legged locomotion remains underexplored.
We propose \textbf{Omni-Perception}, an end-to-end locomotion policy that achieves 3D spatial awareness and omnidirectional collision avoidance by directly processing raw LiDAR point clouds. 
At its core is PD-RiskNet (Proximal-Distal Risk-Aware Hierarchical Network), a novel perception module that interprets spatio-temporal LiDAR data for environmental risk assessment. 
To facilitate efficient policy learning, we develop \textbf{a high-fidelity LiDAR simulation toolkit} with realistic noise modeling and fast raycasting, compatible with platforms such as Isaac Gym, Genesis, and MuJoCo, enabling scalable training and effective sim-to-real transfer.
Learning reactive control policies directly from raw LiDAR data enables the robot to navigate complex environments with static and dynamic obstacles more robustly than approaches relying on intermediate maps or limited sensing. We validate Omni-Perception through real-world experiments and extensive simulation, demonstrating strong omnidirectional avoidance capabilities and superior locomotion performance in highly dynamic environments.
\end{abstract}


\section{Introduction}

Legged robots possess the unique potential to navigate complex, human-centric environments where wheeled or tracked systems often fail.  
Unlocking their full capabilities, however, requires not only agile locomotion but also robust perception and rapid reaction to diverse 3D surroundings \cite{elfes1989using,lee2020learning}.
Achieving safe, high-speed locomotion among static and dynamic obstacles—ranging from ground clutter and uneven terrain to overhanging structures and moving agents—requires holistic spatial awareness and tightly coupled reactive control \cite{he2024agile,xu2023vision}. Enabling such agile, collision-free movement in unstructured 3D environments remains a fundamental challenge \cite{9196777,dudzik2020robust}.

Most state-of-the-art controllers operate blindly using only proprioception, limiting their ability to handle complex terrain or unexpected obstacles \cite{margolis2022walktheseways,margolisyang2022rapid,hwangbo2019learning,lee2020learning,rudin2022learning,kumar2021rma,wang2024arm}. 
While methods incorporating exteroception via depth cameras \cite{cheng2024extreme,kareer2023vinl} have advanced locomotion over uneven terrain, depth-based perception suffers from limitations such as sensitivity to lighting conditions, limited fields of view, noise susceptibility, and the computation overhead of maintaining intermediate representations like elevation maps \cite{agarwal2023legged,plagemann2008learning}. 
These representations, in turn, struggle with aerial clutter and complex non-planar obstacles. 
Alternative approaches that decouple navigation planning from locomotion control \cite{cheng2024navila,miki2022elevation} often lead to conservative behaviors, preventing full exploitation of a robot’s agility.

Meanwhile, LiDAR sensors have become instrumental in fields such as autonomous driving, offering dense, lighting-invariant 3D measurements, and manipulation \cite{li2020lidar,li2020deep,ma2024glover,ma2025glover,ma2024contrastive,zhou2025exploring,zhou2024mitigating,jia2025discoverse}. LiDAR delivers rich geometric information ideal for navigating spatially complex environments. Despite its promise, the integration of LiDAR—\textbf{especially direct point cloud processing}—into end-to-end learning for legged locomotion remains underexplored \cite{bellicoso2018advances,delmerico2019current,rudin2022advanced}. This gap arises from the challenges of real-time point cloud processing within fast control loops, and from difficulties in accurately modeling LiDAR physics for effective sim-to-real transfer \cite{wisth2022vilens}.

This work aims to bridge this critical gap. We posit that directly leveraging raw, spatio-temporal LiDAR point clouds within an end-to-end Reinforcement Learning (RL) policy can unlock robust 3D environmental awareness for legged robots. 
This enables the robot to achieve omnidirectional collision avoidance and agile navigation without relying on handcrafted intermediate representations or restrictive decoupled planning. We introduce \textbf{Omni-Perception}, a novel framework centered around \textbf{an end-to-end policy} trained using RL in a high-fidelity LiDAR simulator. 
Our core technical contribution is the \textbf{Proximal-Distal Risk-Aware Hierarchical Network (PD-RiskNet)}, a novel perception architecture that efficiently processes raw spatio-temporal LiDAR streams to assess multi-level environmental risks. 
To enable scalable and realistic training, we also develop a high-fidelity, cross-platform LiDAR simulation toolkit with realistic noise modeling and efficient parallel raycasting, supporting multiple physics engines including Isaac Gym, Genesis, MuJoCo, and others.
By learning directly from 3D spatial data, Omni-Perception enables legged robots to dynamically track velocity commands while avoiding complex, multi-axis threats (e.g., aerial obstacles, ground traps, moving agents), pushing the boundaries of agile and safe navigation in unstructured environments.Our contributions are summarized as follows:
\begin{enumerate}
    \item \textbf{End-to-End LiDAR-Driven Locomotion Framework:} We present Omni-Perception, the first framework to achieve 3D spatial awareness for legged robots by directly processing raw LiDAR point clouds within an end-to-end RL architecture.
    \item \textbf{Novel LiDAR Perception Network (PD-RiskNet):}
    \item \textbf{High-Fidelity LiDAR Simulation Toolkit:} 
    We develop a high-performance, cross-platform LiDAR simulation toolkit featuring realistic noise models and fast parallel raycasting, enabling effective zero-shot sim-to-real policy transfer.
    \item \textbf{Demonstration of Robust LiDAR-Driven Agility:} We validate Omni-Perception through extensive simulation and real-world experiments, demonstrating strong velocity tracking and omnidirectional avoidance capabilities across diverse and challenging environments containing static and dynamic 3D obstacles.
\end{enumerate}

\section{Related Work}
\label{sec:relatedwork}

\subsection{Learning-Based Legged Locomotion}
Reinforcement Learning (RL) has emerged as a powerful paradigm for developing sophisticated locomotion controllers for legged robots, demonstrating remarkable agility and robustness surpassing traditional model-based methods in many scenarios \cite{hwangbo2019learning,ma2023eureka,zucker2011optimization}. RL approaches have successfully generated controllers for high-speed running \cite{margolisyang2022rapid,kumar2021rma} and challenging terrain traversal \cite{choi2023learning,margolis2022walktheseways}. These policies typically map proprioceptive states—and in some cases, limited exteroceptive information—directly to joint commands. 

\subsection{Exteroceptive Perception for Locomotion}
Incorporating environmental perception is crucial for enabling legged robots to move through unknown or dynamic environments.

\textbf{Depth-Based Perception:} A significant body of work utilizes depth cameras. Many approaches reconstruct the environment into intermediate representations like 2.5D elevation maps, for foothold planning or policy input \cite{hoeller2024anymal,lee2024learning}. However, elevation maps inherently struggle with non-planar obstacles like overhangs or aerial clutter and are sensitive to sensor noise. Other works attempt to use depth data more directly, either by learning end-to-end policies from depth images \cite{zhuang2023robot, zhuang2024humanoid,cheng2024extreme,chanesane2024solo} or extracting sparse features like ray distances from depth \cite{he2024agile}. 
Nevertheless, these methods inherit the limitations of depth sensors, including sensitivity to lighting and limited range. Furthermore, sim-to-real transfer often requires extensive domain randomization, particularly noise injection into depth data, to address real-world sensor characteristics \cite{zhuang2024humanoid}.

\textbf{LiDAR-Based Perception:} LiDAR sensors provide direct 3D measurements and are largely invariant to lighting conditions, making them fundamental in domains such as autonomous driving and mobile robot mapping \cite{miki2022learning,hoeller2021learning,chen2017decentralized}. However, their application within \textit{learning-based, end-to-end legged locomotion} has been very limited. Existing uses often involve traditional pipelines (SLAM, path planning) rather than direct integration into RL policies\cite{wisth2022vilens,bellicoso2018advances,arm2023scientific}. Challenges include the high dimensionality and unstructured nature of point clouds, the computational cost of processing them at high control frequencies, and the difficulty of creating accurate, efficient LiDAR simulations for training \cite{wermelinger2016navigation,yin2024survey}. 
Although some works use simulated LiDAR-like inputs (e.g., ray casting) for perception \cite{he2024agile}, to the best of our knowledge, \textbf{Omni-Perception is the first framework to directly learn end-to-end locomotion policies from raw spatio-temporal LiDAR point clouds}, enabling robust omnidirectional collision avoidance and navigation.

\subsection{Collision Avoidance for Mobile Robots}
Collision avoidance is a fundamental problem in robotics\cite{haddadin2017robot}. Classical approaches often rely on geometric methods and explicit planning in configuration space \cite{gaertner2021collision,park2008uav,lin1998collision}. Model Predictive Control (MPC) has been applied to legged robots, incorporating collision constraints into the optimization \cite{chiu2022collision,lindqvist2020nonlinear}. However, MPC relies on accurate dynamic models, are computationally demanding, and tend to produce slow movements to ensure constraint satisfaction \cite{liao2023walkingnarrowspacessafetycritical,koptev2024reactive}.

Learning-based methods, particularly RL, offer an alternative by enabling robots to learn reactive avoidance behaviors\cite{he2024agile,10904341}. In aerial robotics, RL has been successfully applied to dynamic navigation tasks using perception inputs from RGB-D cameras or simulated laser scans \cite{PerceptionAware2022}. 
Our work advances this line by introducing an end-to-end RL-based collision avoidance strategy that leverages rich 3D information directly from raw LiDAR streams, processed by our specialized PD-RiskNet, integrated tightly within the locomotion control policy.
\section{Method}\label{sec:Method}

\subsection{Problem Formulation}
\label{sec:preliminary} 

We address the problem of learning a continuous locomotion policy $\pi$ for a legged robot, enabling it to track desired velocity commands while performing omnidirectional obstacle avoidance in complex 3D environments using onboard sensing. The policy is trained end-to-end using RL within a high-fidelity simulation environment featuring realistic LiDAR sensing.

The core task is to learn a policy $\pi: \mathcal{O} \rightarrow \Delta(\mathcal{A})$ that maps observations $o_t \in \mathcal{O}$ to a distribution over actions $a_t \in \mathcal{A}$ at each time step $t$. The observation space $\mathcal{O}$ comprises the information available to the policy at time $t$. It includes:
\begin{itemize}
    \item \textbf{Proprioceptive State ($o_t^{\text{prop}}$):} A history of $N_{hist}$ kinematic information, including joint positions ($q_t$), joint velocities ($\dot{q}_t$), base linear and angular velocities ($v_t, \omega_t$), and base orientation represented by the projected gravity $g_t$. 
    \item \textbf{Exteroceptive State ($P_t$):} A history of $N_{hist}$ raw 3D point cloud data, providing spatio-temporal information about the surrounding environment.
    \item \textbf{Task Command ($c_t$):} The desired velocity command, typically a target linear velocity $v_t^{\text{cmd}}$ and angular velocity $\omega_t^{\text{cmd}}$ in the robot's base frame.
\end{itemize}
Thus, the observation is $o_t = (o_{t-N_{hist}+1:t}^{\text{prop}}, P_{t-N_{hist}+1:t}, c_t)$. Our proposed perception network, PD-RiskNet (Sec.~\ref{sssec:pdrsknet}), processes the sequence of point clouds $P_{t-N_{hist}+1:t}$ and integrates the extracted features with the proprioceptive history $o_{t-N_{hist}+1:t}^{\text{prop}}$ and command $c_t$ to inform the policy.

The action space $\mathcal{A}$ consists of low-level motor commands $a_t$, specifically the target joint positions issued to the robot’s actuators. We frame the learning problem as optimizing the policy within an infinite-horizon discounted Markov Decision Process (MDP), defined by the tuple $(\mathcal{S}, \mathcal{A}, r, \gamma, T)$. While the policy directly uses observations $o_t$, the underlying state $s_t \in \mathcal{S}$ may potentially include hidden simulator states or history. The transition dynamics $T: \mathcal{S} \times \mathcal{A} \rightarrow \Delta(\mathcal{S})$ are governed by the physics simulator and sensor models. The reward function $r: \mathcal{S} \times \mathcal{A} \rightarrow \mathbb{R}$ (or approximated as $r(o_t, a_t)$) is designed to encourage specific behaviors (detailed in Sec.~\ref{sec:policy_learning}). 
The objective is to find the optimal policy $\pi^*$ that maximizes the expected discounted sum of future rewards:
\begin{equation}
\pi^* = \arg\max_{\pi} \mathbb{E}_{\tau \sim \pi} \left[ \sum_{t=0}^{\infty} \gamma^t r(s_t, a_t) \right],
\end{equation}
where $\tau = (s_0, a_0, s_1, a_1, \dots)$ is a trajectory generated under policy $\pi$, $a_t \sim \pi(o_t)$, and $\gamma \in [0, 1)$ is the discount factor.

\subsection{Custom LiDAR Rendering Framework} 
\label{subsec:custom_lidar}
\paragraph{Motivation and Backend.}
While existing simulators, like NVIDIA Isaac Gym/Sim~\cite{makoviychuk2021isaac}, Gazebo~\cite{1389727}, Genesis~\cite{Genesis}, provide physics simulation capabilities, their rendering solutions often lack support for diverse lidar sensor models like Non-repetitive scan lidar or impose limitations on parallelism and cross-platform execution. To address these gaps, inspired by ~\cite{kulkarni2025aerial}, we introduce a custom, high-performance LiDAR rendering framework that leverages NVIDIA Warp~\cite{warp2022} for GPU acceleration and Taichi~\cite{hu2019taichi} for cross-platform compatibility, enabling execution on systems lacking dedicated NVIDIA GPUs, including CPU-only environments.

\paragraph{Optimized Mesh Management for Parallel Simulations.}
Simulating numerous environments with frequently moving dynamic objects poses a significant challenge. Transforming each mesh and rebuilding acceleration structures (BVHs) for each environment independently at each timestep~\cite{lauterbach2009fast} become computationally prohibitive in such scenarios. To overcome this bottleneck, we employ a specialized mesh partitioning and update strategy:
\begin{itemize}
    \item \textbf{Per-Environment Static Mesh:} For each environment $\boldsymbol{\mathcal{E}}_i$, non-moving geometry (terrain, fixed obstacles) is represented as a static mesh $\boldsymbol{\mathcal{M}}^{\textrm{static}}_i$. Its BVH structure is built only once upon initialization, minimizing redundant computations.
    \item \textbf{Global Shared Dynamic Mesh:} We aggregate the meshes of \textit{all} dynamic entities  (e.g., robot parts, moving obstacles) from \textit{all} $N_{\text{envs}}$ parallel environments into a single, global dynamic mesh $\boldsymbol{\mathcal{M}}^{\textrm{dynamic}}_{\textrm{global}, t}$. At simulation time $t$, only the vertex positions of this shared mesh are updated based on the latest transformations ($\mathbf{T}_{j,t}$) of the corresponding dynamic objects $\boldsymbol{\mathcal{M}}^{\textrm{dynamic}}_j$ across all environments.
\end{itemize}

\paragraph{Synchronous Update and Raycasting.}
LiDAR simulation involves raycasting against both the appropriate static mesh ($\boldsymbol{\mathcal{M}}^{\textrm{static}}_i$) and the \textit{single} global dynamic mesh ($\boldsymbol{\mathcal{M}}^{\textrm{dynamic}}_{\textrm{global}, t}$). Crucially, the vertex update for $\boldsymbol{\mathcal{M}}^{\textrm{dynamic}}_{\textrm{global}, t}$ occurs synchronously for all environments, typically matching the sensor's update frequency. This centralized update mechanism avoids the substantial overhead of individually updating dynamic meshes and rebuilding their acceleration structures for each of the $N_{\text{envs}}$ environments at every step. 
This enables all parallel environments to efficiently query the same up-to-date dynamic geometry, resulting in scalable, high-performance simulation across diverse hardware configurations.

\paragraph{LiDAR Model Support and Scan Pattern Simulation} 
Leveraging this framework, we have implemented support for a wide range of commercial LiDAR models. This includes non-repetitive scanning LiDARs, such as the Livox series (e.g., Mid-360, Avia), where we simulate their characteristic scan patterns by utilizing real device data to determine scan angles synchronized with the simulation time $t$. Furthermore, support extends to conventional rotating LiDARs, with pre-configured models including the Velodyne series (e.g., HDL-64, VLP-32) and various Ouster sensors.

\subsection{Omni-Perception Framework}
\begin{figure}[t!]
\centering
\includegraphics[width=1\linewidth]{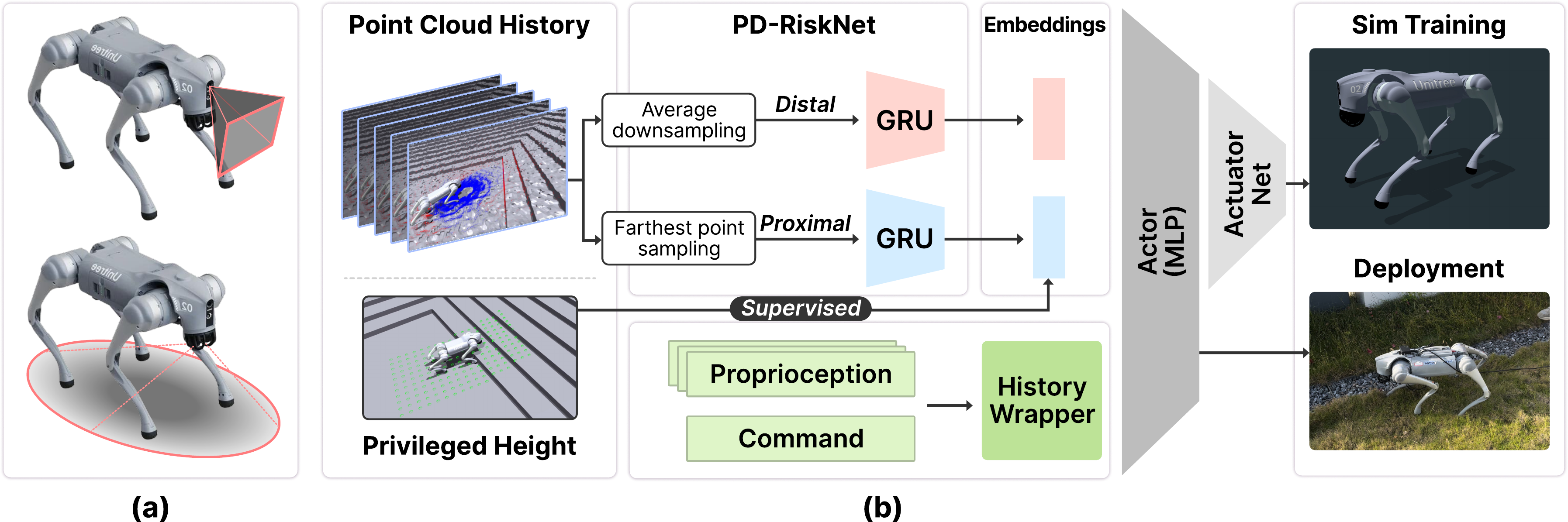}
\vspace{-0.2in}
\caption{\small Proposed System Framework. (a) Visualization of differing sensor coverage: the typically narrow, forward-directed field of view of a depth camera (top) contrasted with the broader and longer range, coverage of a LiDAR sensor (bottom), shown on the Unitree Go2 robot. (b) Detailed diagram of the perception and control pipeline. Raw point cloud history is processed via two pathways within PD-RiskNet (average downsampling for distal features, farthest point sampling for proximal features, each fed to a GRU) generating spatial embeddings. In the point cloud history visualization, the dense blue points represent the proximal point cloud, while the sparse red points depict the distal point cloud. These embeddings are concatenated with historical proprioception and commands processed by a History Wrapper, and supervised using privileged height information. The combined features are input to an Actor (MLP) network that outputs actuator targets for sim-to-real training and deployment.}
\label{fig:framework}
\vspace{-0.2in}
\end{figure}

Our framework, shown in Figure~\ref{fig:framework}, consists of the PD-RiskNet for processing LiDAR data and a locomotion policy that integrates this perception with proprioception and commands.

\subsubsection{PD-RiskNet: Processing Spatio-Temporal Point Clouds}
\label{sssec:pdrsknet}

The PD-RiskNet architecture is designed to process spatio-temporal point cloud data acquired from a legged robot's LiDAR sensor. The initial step involves partitioning the raw point cloud $P_{raw}$ into two distinct subsets: the proximal point cloud $P_{proximal}$ and the distal point cloud $P_{distal}$. 
This partitioning is based on a vertical angle threshold $\theta$, distinguishing near-field points (higher $\theta$) from far-field points (lower $\theta$), effectively separating dense local geometry from sparse distant observations.

\textbf{Proximal Point Cloud Processing:} $P_{proximal}$, representing the near-field environment, is generally characterized by higher point density and relatively smaller variations in range. To efficiently process this dense data while preserving crucial local geometric details, Farthest Point Sampling (FPS)\cite{qi2017pointnetplusplus} is employed to obtain a sparse yet representative subset of points. Following sampling, these points are structured by sorting based on their \textbf{spherical coordinates} ($\theta$ then $\phi$). This ordered representation of the sampled proximal point cloud is then input to a dedicated Gated Recurrent Unit (GRU). The feature extraction capability of this GRU is enhanced during training using a Privileged Height signal as supervision, guiding the network to learn representations specifically pertinent to local terrain properties and potential risks for legged locomotion.

\textbf{Distal Point Cloud Processing:} Conversely, $P_{distal}$, covering the far-field environment, is typically sparse and exhibits larger variations in range values. To handle the data characteristics and reduce the influence of outliers in this sparse data, Average Downsampling is applied. This operation yields a more uniform representation of distant structures. 
Downsampled distal point clouds from the current and preceding $N_{hist}$ frames are combined, forming a spatio-temporal sequence. 
This sequence, structured by sorting points based on their spherical coordinates ($\theta$ then $\phi$) after downsampling, is processed by a separate GRU module to extract features capturing the dynamic environmental context at range.

\subsubsection{Risk-Aware Locomotion Policy}
\label{sec:policy_learning}

The locomotion policy is implemented as an MLP that takes the concatenated features from PD-RiskNet and the proprioceptive history as input.

\textbf{Observation Space.} As detailed in Sec.~\ref{sec:preliminary}, the observation includes proprioceptive history ($o_{t-N_{hist}+1:t}^{\text{prop}}$), processed LiDAR features from PD-RiskNet ($f_t^{PD}$) derived from $P_{t-N_{hist}+1:t}$, and the current velocity command ($c_t$). Hence, the policy input is represented as $o_t^{policy} = (o_{t-N_{hist}+1:t}^{\text{prop}}, f_t^{PD}, c_t)$.

\textbf{Action Space.} The policy outputs target joint positions $a_t$ for a low-level PD controller.

\textbf{Reward Function.} The reward function $r_t = r(s_t, a_t)$ is designed to train the robot to follow velocity commands while actively avoiding collisions by leveraging LiDAR data. We use a similar reward function as \cite{cheng2024extreme}. A complete list of reward functions can be found in Appendix.~\ref{appendix:rewards_omni}. It primarily consists of two novel components focused on velocity tracking with integrated avoidance and maximizing environmental clearance:
 \textbf{(1) Linear Velocity Tracking with Avoidance ($r_{vel\_avoid}$):} This encourages tracking a modified target linear velocity that combines the external command $v_t^{cmd}$ with a dynamically computed avoidance velocity $V_t^{\text{avoid}}$. 
\begin{wraptable}{r}{0.4\textwidth}
\vspace{-15pt}
  \centering
    \includegraphics[width=0.4\textwidth]{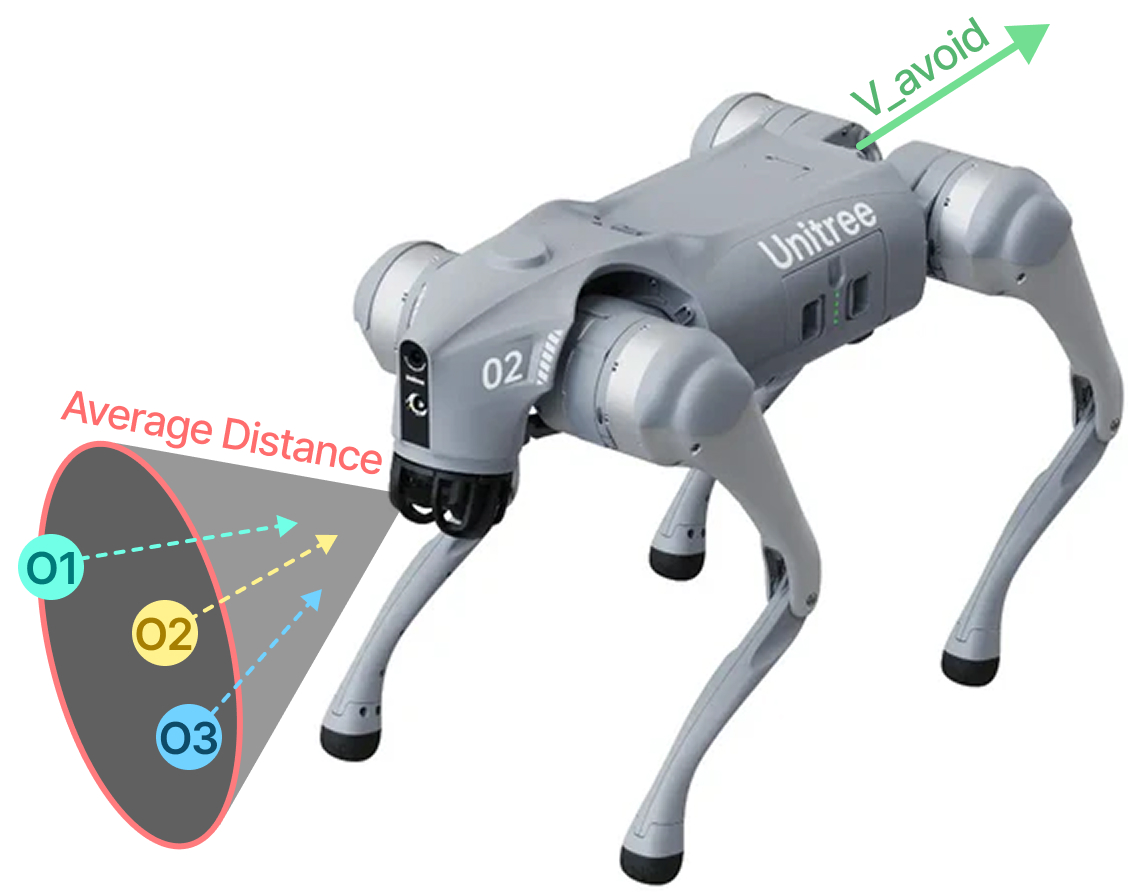}
   \captionsetup{type=figure}
   \caption{The calculation of the sector-based avoidance velocity.
   \vspace{5pt}
   }
   \label{fig:v_caculation}
\end{wraptable}
The avoidance velocity $V_t^{\text{avoid}}$ actively pushes the robot away from nearby obstacles detected by LiDAR. The robot's 360° horizontal surroundings are divided into $N_{sec}=36$ angular sectors. Within each sector $j$, the minimum obstacle distance $d_t^{j}$ is found. If $d_t^{j}$ is below a threshold $d^{thresh}$ (e.g., 1m), an avoidance velocity component $V_{t,j}^{avoid}$ is generated, pointing away from the sector $j$. Its magnitude is calculated based on proximity:  
        $$
        \|V_{t,j}^{avoid}\| = \exp(-d_{j,t} \cdot \alpha_{avoid})
        $$
The total avoidance velocity $V_t^{\text{avoid}} = \sum_{j=1}^{N_{sec}} V_{t,j}^{avoid}$ is the vector sum over all sectors. The reward term penalizes deviation from this combined target velocity:
        $$
        r_{vel\_avoid}= \exp(-\beta_{va}*\|v_t - (v_t^{cmd} + V_{avoid,t})\|^2)
        $$where $v_t$ is the base linear velocity, and $\beta_{va}$ is weighting coefficients. 
\textbf{(2) Distance Maximization ($r_{rays}$):} This reward encourages maintaining a safety margin and pushing towards open areas by maximizing the capped LiDAR ray distances. Using the capped distances $\hat{d}_{t,i} = \min(d_{t,i}, d_{\text{max}})$ from $n$ representative distal LiDAR rays, the reward is:
        $$
        r_{rays} =\sum_{i=1}^{n} \frac{\hat{d}_{t,i}}{n \cdot d_{\text{max}}} \label{eq:reward_rays_simplified}
        $$

\textbf{Domain Randomization.} This was applied during training to enhance sim-to-real transfer. This included randomizing robot physical parameters \cite{margolis2022walktheseways} and LiDAR perception characteristics (masking, noise). See Appendix \ref{appendix:dr_params} for details.
\section{Experiments}
\label{sec:experiments}


\subsection{LiDAR fidelity Evaluation}
\label{sec:lidar_sim}
\begin{figure}[htbp!]
\centering
\vspace{-0.1in}
\setlength{\abovecaptionskip}{0cm}
\setlength{\belowcaptionskip}{0cm}
\includegraphics[width=1\linewidth]{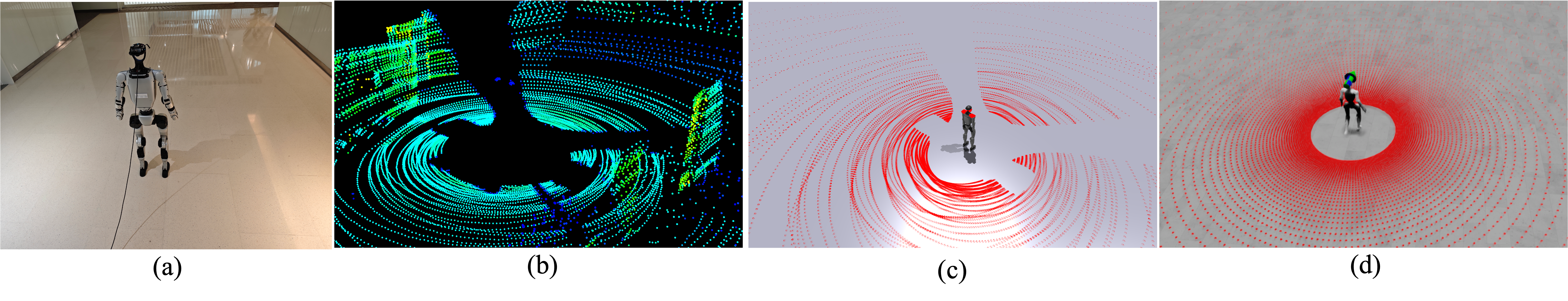}
\vspace{-0.2in}
\caption{\small Comparison of simulated and real point cloud for the Unitree G1 robot. (a) The physical Unitree G1 robot setup. (b) Real-world LiDAR scan captured by the onboard Livox Mid-360 sensor. (c) \textbf{(Ours)} Point cloud generated using our Livox Mid-360 sensor model within the Isaac Gym. (d) Point cloud using the official sensor within the Isaac Sim. Ours captures the self-occlusion effect as in the real-world LiDAR.}.
\label{fig:Lidar_simulation_real}
\end{figure}

As illustrated in Figure.~\ref{fig:Lidar_simulation_real}, we validate our LiDAR simulator's fidelity. We compare its output to real data from a Unitree G1 robot using a Livox Mid-360 sensor and Isaac Sim LiDAR using the same horizontal angle and vertical angle. Our simulation includes the sensor's non-repetitive scan pattern and robot self-occlusion. The simulated and real point distributions and structures appeared similar. This result indicates high simulator fidelity, which helps reduce the sim2real perception gap for LiDAR policies.You can find more lidar model scan patterns in the Appendix.\ref{appendix:Lidar_pattern}

\begin{table}[htbp!]
  \centering
  \setlength{\abovecaptionskip}{0cm}
\setlength{\belowcaptionskip}{0cm}
  \small 
  \setlength{\tabcolsep}{3pt} 
  \vspace{-1.3em} 
  \caption{\small Rendering time (ms) for static scenes across configurations. Ours is much more efficient than Isaac Sim.}
  \label{tab:efficiency}

  \scalebox{0.8}{
  \begin{tabular}{@{}l *{10}{S[table-format=4.1]}@{}}
    \toprule
    \multirow{3}{*}{\textbf{Lines}} & \multicolumn{10}{c}{\textbf{Number of Environments}} \\
    \cmidrule(lr){2-11}
    & \multicolumn{2}{c}{\textbf{256}} & \multicolumn{2}{c}{\textbf{512}} & \multicolumn{2}{c}{\textbf{1024}} & \multicolumn{2}{c}{\textbf{2048}} & \multicolumn{2}{c}{\textbf{4096}} \\
    \cmidrule(lr){2-3} \cmidrule(lr){4-5} \cmidrule(lr){6-7} \cmidrule(lr){8-9} \cmidrule(lr){10-11}
    & \multicolumn{1}{c}{Ours} & \multicolumn{1}{c}{IsaacSim}
    & \multicolumn{1}{c}{Ours} & \multicolumn{1}{c}{IsaacSim}
    & \multicolumn{1}{c}{Ours} & \multicolumn{1}{c}{IsaacSim}
    & \multicolumn{1}{c}{Ours} & \multicolumn{1}{c}{IsaacSim}
    & \multicolumn{1}{c}{Ours} & \multicolumn{1}{c}{IsaacSim} \\
    \midrule
    1k & {\textbf{2.5$_{\pm0.3}$}} & {21$_{\pm0.7}$} & {\textbf{2.1$_{\pm0.4}$}} & {19$_{\pm1.7}$} & {\textbf{2.2$_{\pm0.1}$}} & {32$_{\pm0.8}$} & {\textbf{3.4$_{\pm0.2}$}} & {40$_{\pm1.8}$} & {\textbf{3.3$_{\pm0.3}$}} & {95$_{\pm2.1}$} \\
    4k & {\textbf{3.3$_{\pm0.4}$}} & {28$_{\pm0.6}$} & {\textbf{5.7$_{\pm0.3}$}} & {39$_{\pm0.3}$} & {\textbf{16.2$_{\pm1.1}$}} & {81$_{\pm1.1}$} & {\textbf{49.9$_{\pm2.3}$}} & {146$_{\pm4.8}$} & {\textbf{117.9$_{\pm0.7}$}} & {308$_{\pm5.8}$} \\
    16k & {\textbf{3.7$_{\pm0.1}$}} & {81$_{\pm3.1}$} & {\textbf{11.2$_{\pm0.8}$}} & {142$_{\pm1.4}$} & {\textbf{30.4$_{\pm2.8}$}} & {317$_{\pm17.3}$} & {\textbf{118.3$_{\pm4.7}$}} & {600$_{\pm8.8}$} & {\textbf{220.0$_{\pm1.1}$}} & \multicolumn{1}{c}{OOM} \\ 
    32k & {\textbf{5.8$_{\pm0.1}$}} & {148$_{\pm2.3}$} & {\textbf{22.9$_{\pm2.6}$}} & {286$_{\pm2.7}$} & {\textbf{47.9$_{\pm0.8}$}} & {569$_{\pm14.9}$} & {\textbf{133.5$_{\pm8.4}$}} & {1133$_{\pm32.3}$} & {\textbf{273.0$_{\pm0.9}$}} & \multicolumn{1}{c}{OOM} \\ 
    \bottomrule
  \end{tabular}
  }
  \vspace{-0.3em} 
\end{table}

\subsection{Comparison with Existing Simulators and Efficiency Analysis}
\label{subsec:simulator_comparison}

While numerous robotics simulators exist, official support for LiDAR simulation is relatively 
\begin{wraptable}{r}{0.6\textwidth}
    \centering
    \vspace{-1em} 
    \caption{\small Comparison of Capabilities Across Platforms.}
    \scalebox{0.8}{ 
        \begin{tabular}{lccc}
            \toprule
            \textbf{LiDAR Simulation Feature} & \textbf{Ours} & \textbf{Isaac Sim} & \textbf{Gazebo} \\
            \midrule
            Rotating LiDAR Support           & \checkmark & \checkmark & \checkmark \\
            Solid-State LiDAR Support        & \checkmark & \checkmark & \checkmark \\
            Hybrid Solid-State LiDAR         & \checkmark & \poor & \checkmark \\ 
            \quad \textit{(Non-repetitive scan)} & & & \\
            Static Irregular Objects         & \checkmark & \checkmark & \checkmark \\
            Dynamic Irregular Objects        & \checkmark & \poor & \poor \\ 
            Self-Occlusion                   & \checkmark & \poor & \poor \\
            Cross-Platform Support           & \checkmark & \poor & \checkmark \\ 
            Massively Parallel Execution   & \checkmark & \checkmark & \poor \\ 
            \bottomrule
        \end{tabular}
    } 
    
    \label{tab:simulator_features}
    \vspace{-2em} 
\end{wraptable}
limited, with Gazebo and NVIDIA Isaac Sim being prominent examples offering built-in capabilities.  We compare our LiDAR toolkit features against these established platforms based on our evaluation and publicly available documentation. Table~\ref{tab:simulator_features} summarizes the key LiDAR simulation capabilities across these platforms. Table~\ref{tab:efficiency} shows the rendering speed of our liDAR implementation relative to Isaac Sim. Our implementation aims to combine broad feature support with a high-performance parallel execution environment, like Isaac Gym and Genesis.You can find more detail in the Appendix.~\ref{appendix:Lidar_Efficiency}.
\subsection{PD-RiskNet Ablation Study}
We have tested four ways to process the LiDAR point cloud to see if our PD-RiskNet is effective.
\begin{wraptable}{r}{0.5\textwidth}
\centering
\caption{\small PD-RiskNet Ablation Results (30 Trials)}
\label{tab:ablation_simple}
\scalebox{0.8}{
\begin{tabular}{@{}lcc@{}}
\toprule
\textbf{Method} & \textbf{Success Rate (\%)} & \textbf{Collision Rate (\%)} \\
\midrule
Direct MLP & OOM* & OOM* \\
FPS + MLP & 33.3\% & 93.3\% \\
FPS + GRU & 30.0\% & 70.0\% \\
\midrule
\textbf{Ours} & \textbf{76.7\%} & \textbf{56.7\%} \\
\bottomrule
\end{tabular}
}
\vspace{-1em} 
\end{wraptable} 
We compare four approaches for extracting the same dimensional point cloud features in an Isaac Gym simulation with 6 dynamic obstacles (1.5 m/s speed) over 30 trials each. As shown in Fig.~\ref{fig:sim_exp}, the task is to reach the end of a 12m x 4m area. 
 Table.~\ref{tab:ablation_simple} shows the direct method using an MLP failed due to Out of Memory(OOM) (24GB). Methods using FPS downsampling, FPS+MLP, and FPS+GRU suffer very low success rates, mostly failing due to timeouts or getting stuck rather than colliding.

\subsection{Real-World Deployment}
\begin{figure}[htbp]
  \centering 
\vspace{-8pt}  
  \begin{minipage}[t]{0.48\textwidth}
    \centering
    \includegraphics[width=\linewidth, valign=t]{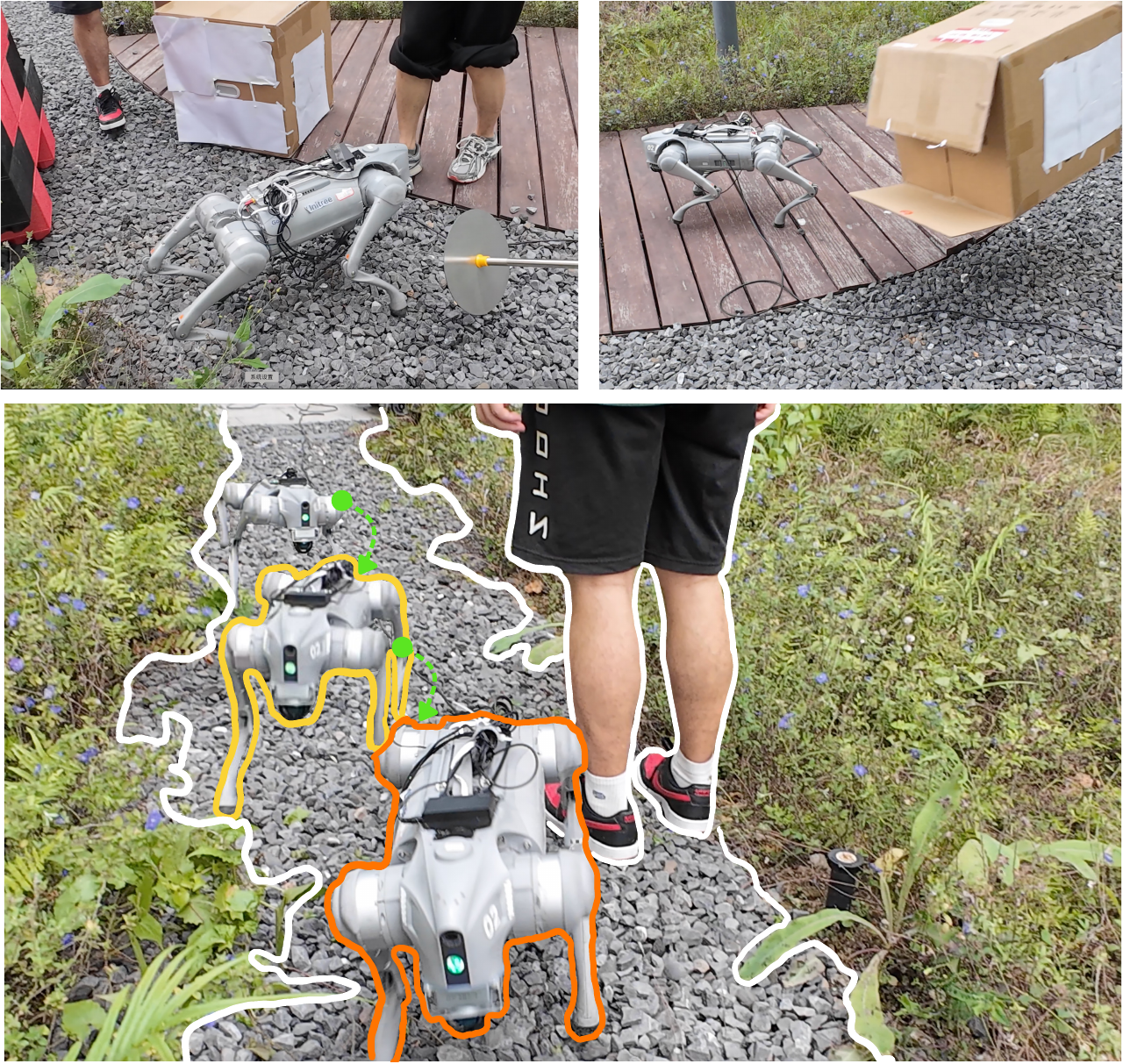}
    \caption{\small Robot obstacle avoidance performance was assessed in varied scenarios, including complex terrain and dynamic human interference.} 
    \label{fig:real_exp}
  \end{minipage}
  \hfill 
  \begin{minipage}[t]{0.48\textwidth}

    \begin{minipage}[t]{\linewidth} 
        \centering
        \includegraphics[width=\linewidth, valign=t]{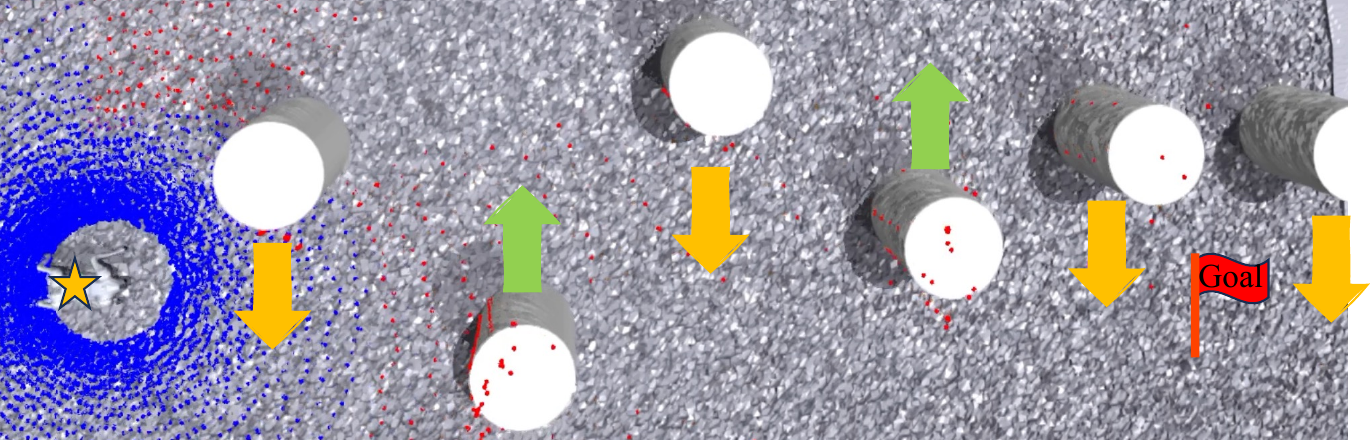}
        \caption{\small Simulation ablation experiment scene setting.} 
        \label{fig:sim_exp}
    \end{minipage}

    \par\vspace{\medskipamount} 

    \begin{minipage}[b]{\linewidth} 
        \centering
        \renewcommand{\arraystretch}{1} 
        \resizebox{\linewidth}{!}{%
          \begin{tabular}{lcc}
            \toprule
            \textbf{Scenario} & \textbf{Omni-Perception} & \textbf{Unitree System} \\
            \midrule
            Static obstacles & \textbf{30/30 (100\%)} & 30/30 (100\%) \\
            Aerial obstacles & \textbf{21/30 (70\%)} & 0/30 (0\%) \\
            Small obstacles & 25/30 (83\%) & \textbf{30/30 (100\%)} \\
            Moving humans & \textbf{27/30 (90\%)} & 0/30 (0\%) \\
            \bottomrule
          \end{tabular}%
        } 
        \captionof{table}{\small Real-world performance comparison between Omni-Perception and the native Unitree system. Success rates are reported based on 30 trials for each scenario involving different types of obstacles. Bold values indicate the higher success rate for that scenario.}
        \label{tab:real_test}
    \end{minipage}

  \end{minipage} 

\vspace{-20pt} 
\end{figure}

We evaluate the omnidirectional obstacle avoidance effectiveness and its reactive behaviors in various scenarios, as shown in Fig.~\ref{fig:firstPage}. Furthermore, we test its performance under several extreme conditions. As depicted in Fig.~\ref{fig:real_exp}, the robot successfully crosses through rocky terrain within dense grass, even while experiencing interference from moving humans. It also demonstrates adaptability to rapidly approaching aerial obstacles and frequent human obstructions. A quantitative comparison of success rates against the native Unitree system across different obstacle types over 30 trials is presented in Table~\ref{tab:real_test}.
\section{Conclusion} 
In conclusion, we present Omni-Perception, an end-to-end reinforcement learning framework that successfully enables robust, omnidirectional collision avoidance for legged robots by directly leveraging raw LiDAR point cloud data. Through PD-RiskNet architecture and training in a high-fidelity lidar simulation environment, our approach integrates perception and control, allowing robots to move in complex environments with diverse static and dynamic obstacles while maintaining desired velocity commands. Real-world experiments validated the system's ability to perform agile and safe locomotion in challenging, dynamic 3D settings.
\newpage
\section{Limitations}

\textbf{Environmental Geometry:} Performance can degrade in environments dominated by highly unstructured vegetation, such as dense grass, where LiDAR struggles to extract reliable geometric features essential for locomotion. This poses a challenge similar to sensor limitations noted in visually complex scenes.Introducing semantic segmentation into the framework is a potential solution.

\textbf{Sim2Real Fidelity Trade-off:} the current approach utilizes sampling strategies within a high-fidelity simulator to facilitate sim-to-real transfer. While effective, this involves an inherent trade-off, potentially reducing the level of fine-grained geometric detail perceived by the robot compared to the raw sensor output. This simplification might become a limiting factor in scenarios demanding exceptionally precise navigation around complex small obstacles, potentially leading to suboptimal paths or collisions. Further research could focus on improving simulation fidelity, developing more advanced domain randomization or adaptation techniques, or incorporating online learning to fine-tune perception upon deployment.


\subsection{Failure Case}
\textbf{extremely unstructured environments:} Although we have successful examples in this environment, the effectiveness of the robot's locomotion strategy will be greatly reduced. The robot identified the grass on the left side of the picture as a dangerous obstacle, and because it was close, it quickly moved to the left. 

\begin{figure}[htbp!]
\vspace{0pt}
  \centering
    \includegraphics[width=0.4\textwidth]{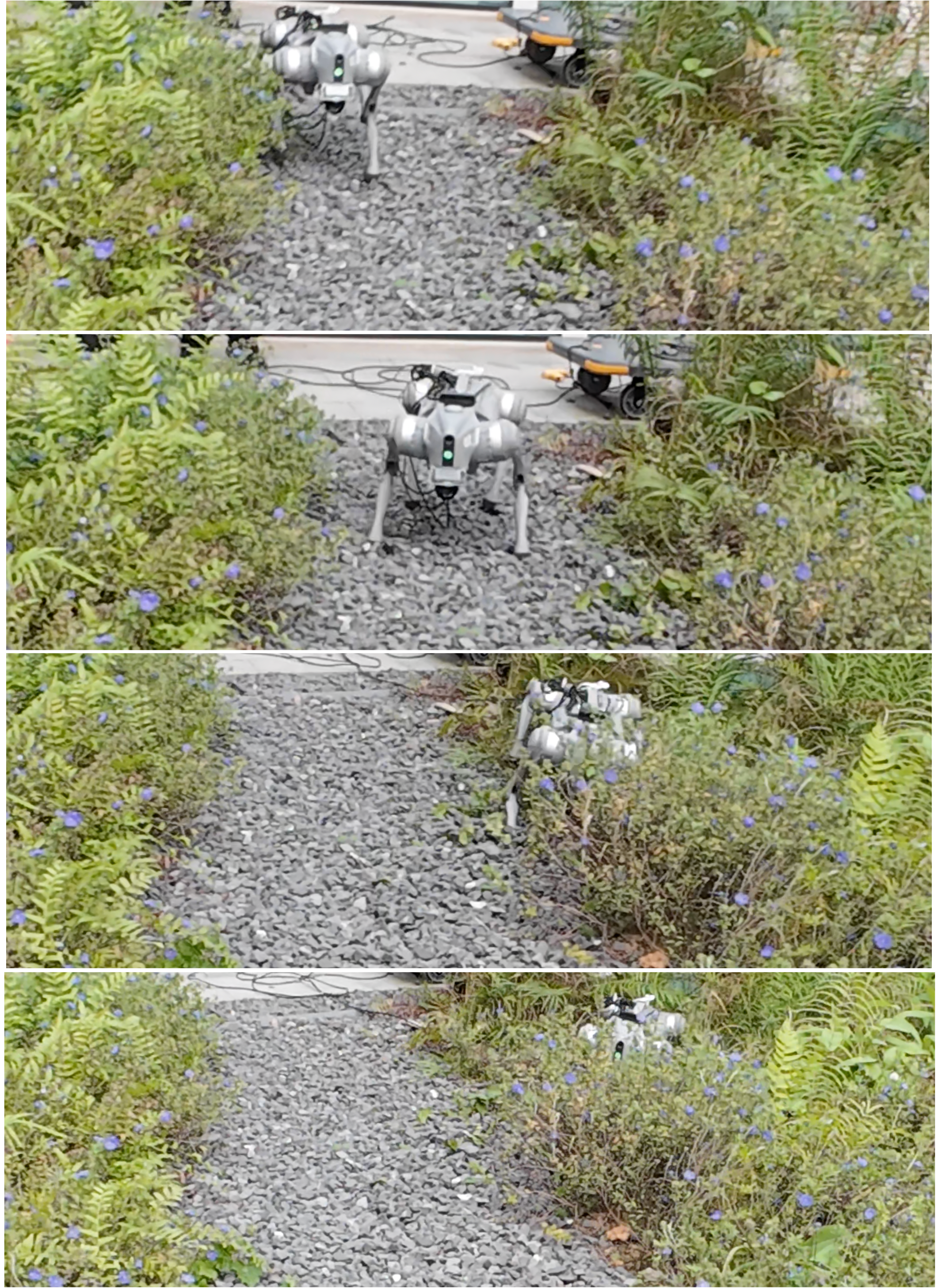}
   \captionsetup{type=figure}
   \caption{\small dense grass.}
   \label{fig:v_caculation}
\end{figure}

Because the entire passage was narrow, the robot was forced to enter the grass on the right side of the picture. After entering the grass, the robot's surroundings were perceived as a dangerous area, which led to mission failure.

\textbf{Objects that are too small and sparse:} 

\begin{wraptable}{r}{0.4\textwidth}
\vspace{-15pt}
  \centering
    \includegraphics[width=0.4\textwidth]{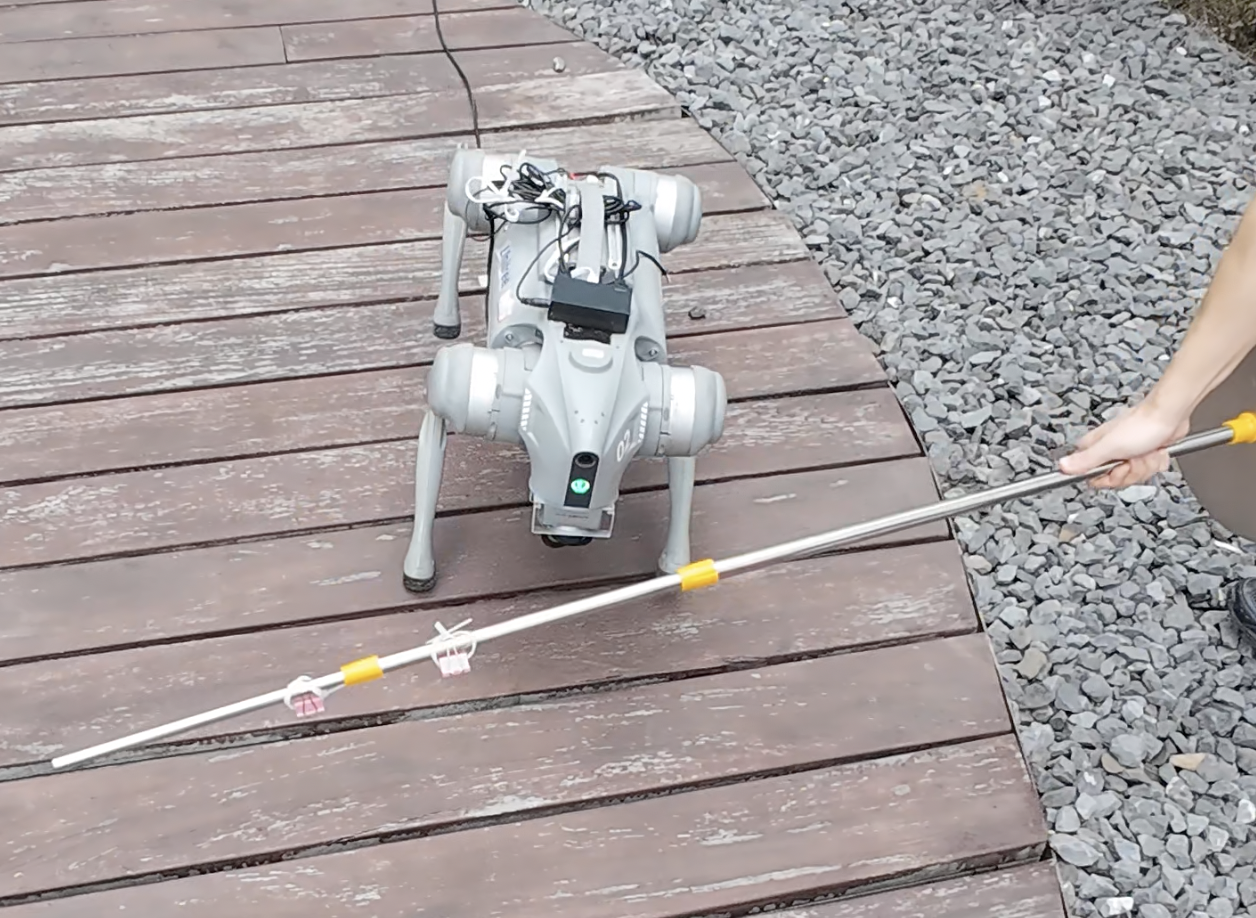}
   \captionsetup{type=figure}
   \vspace{-15pt}
   \caption{\small Thin rod.}
   \label{fig:v_caculation}
\end{wraptable}
Since we will average the distant point clouds, the features of very small objects will be destroyed, making it impossible to respond correctly to such obstacles.

\clearpage
\acknowledgments{This work was supported by the National Natural Science Foundation of China (No. 62306257) and the Guangzhou-HKUST(GZ) Joint Funding Program (Grant No.2023A03J0008), Education Bureau of Guangzhou Municipality.} 


\bibliography{example} 

\newpage
\appendix

\section{Implementation Details}
\subsection{Rewards}

\begin{table}[htbp!]
  \centering 
    {
    \scriptsize 
    \begin{tabular}{@{}llr@{}} 
    \toprule
    Term & Equation & Weight \\ [0.5ex]
    \midrule
    \multicolumn{3}{l}{\textit{\textcolor{green}{Omni-Perception Rewards}}} \\ 
    Velocity Tracking with Avoidance ($r_{vel,avoid}$) & \( \exp\{{-\beta_{va} ||\mathbf{v}_{t} - (\mathbf{v}_{t}^{\text{cmd}} + \mathbf{V}_{\text{avoid},t})||^2}\} \) & \textit{2} \\ [1ex] 
    Distance Maximization ($r_{rays}$) & \( \sum_{i=1}^{n} \frac{\min(d_{t,i}, d_{\max})}{n \cdot d_{\max}} \) & \textit{1.5} \\ [1ex] 
    \midrule
    \multicolumn{3}{l}{\textit{\textcolor{red}{Auxiliary Rewards}}} \\ 
    z velocity & \(v_{z}^2\) & \num{-3e-4} \\ [0.5ex]
    foot stumble & \(||Force^{\text{foot}}_{xy}||^2\) & \num{-2e-2} \\ [0.5ex]
    link collision &\(||Force^{\text{PenltyLink}}_{xy}||^2\)  & $-0.02$ \\ [0.5ex] 
    joint limit violation & \( \mathbf{1}_{q_i>q_{\max} || q_i < q_{\min}}\) & $-0.2$ \\ [0.8ex] 
    joint torques & \(||\boldsymbol{\tau}||^2\) & \num{-1e-6} \\ [0.5ex]
    joint velocities & \(||\dot{\mathbf{q}}||^2\) & \num{-1e-6} \\ [0.5ex]
    joint accelerations & \(||\ddot{\mathbf{q}}||^2\) & \num{-2.5e-7} \\ [0.5ex]
    action smoothing & \(||\mathbf{a}_{t-1} - \mathbf{a}_t||^2\) & \num{-5e-3} \\ [0.5ex]
    action smoothing rate & \(||\mathbf{a}_{t-2} - 2\mathbf{a}_{t-1} + \mathbf{a}_t||^2\) & \num{-5e-3} \\ [0.5ex]
    \bottomrule
    \end{tabular}
    }
    \caption{Reward structure for Omni-Perception}
    \label{appendix:rewards_omni}
\end{table}

\section{Network Architecture Details}
\label{sec:appendix_network}

The Omni-Perception framework utilizes specific neural network architectures for perception (PD-RiskNet) and control (Actor).

\subsection{PD-RiskNet Architecture}

The PD-RiskNet processes spatio-temporal LiDAR point cloud data. As described in Section 3.3.1 of the main text, the raw point cloud is partitioned into proximal ($P_{\text{proximal}}$) and distal ($P_{\text{distal}}$) subsets.

\begin{itemize}
    \item \textbf{Input Processing:} Both the proximal and distal pathways process a history of point cloud data. The paper mentions using a history of $N_{\text{hist}}$ frames (Sec 3.1, Sec 3.3.1). Based on additional details provided, we use $N_{\text{hist}}=10$, meaning each Gated Recurrent Unit (GRU) processes features derived from 10 consecutive LiDAR scans.
    \item \textbf{Proximal Pathway:} The proximal point cloud ($P_{\text{proximal}}$) undergoes Farthest Point Sampling (FPS) before being fed into a dedicated GRU. This GRU is supervised using privileged height information during training. The output embedding dimension from the proximal GRU is \textbf{187 features}.
    \item \textbf{Distal Pathway:} The distal point cloud ($P_{\text{distal}}$) is processed using Average Downsampling. Features from the current and $N_{\text{hist}}-1$ preceding frames (forming a sequence of 10 frames) are fed into a separate GRU. The output embedding dimension from the distal GRU is \textbf{64 features}.
\end{itemize}

The embeddings from both the proximal and distal GRUs are concatenated with proprioceptive history and the command vector before being passed to the Actor network.

\subsection{Actor Network Architecture}

The locomotion policy (Actor) is implemented as a Multi-Layer Perceptron (MLP).

\begin{itemize}
    \item \textbf{Input:} Concatenated features from PD-RiskNet (187 + 64 features), processed proprioceptive history, and the current velocity command.
    \item \textbf{Hidden Layers:} The MLP consists of sequential fully connected layers with the following hidden dimensions: \textbf{[1024, 512, 256, 128]}. Appropriate activation functions ELU are typically used between layers.
    \item \textbf{Output Layer:} The final layer outputs the target joint positions for the robot's actuators. The output dimension is \textbf{12}.
\end{itemize}

\section{PPO Hyperparameters}
\label{sec:appendix_ppo}

The policy was trained using the Proximal Policy Optimization (PPO) algorithm \cite{schulman2017proximal}. Key hyperparameters used during training are listed in Table \ref{tab:ppo_params}.

\begin{table}[h!] 
\centering
\caption{PPO Hyperparameters}
\label{tab:ppo_params}
\scalebox{0.8}{
\begin{tabular}{@{}lc@{}} 
\toprule
Parameter                          & Value \\ \midrule
PPO clip parameter ($\epsilon$)    & 0.2   \\
GAE $\lambda$                      & 0.95  \\
Reward discount factor ($\gamma$)  & 0.99  \\
Learning rate                      & \num{1e-3}  \\ 
Learning rate schedule             & adaptive \\ 
Value loss coefficient             & 1.0   \\ 
Use clipped value loss             & True  \\ 
Entropy coefficient                & 0.01  \\ 
Desired KL divergence              & 0.01  \\ 
Max gradient norm                  & 1.0   \\ 
Number of environments             & 4096  \\ 
Number of env steps per batch      & 24    \\ 
Learning epochs per batch          & 5     \\ 
Number of mini-batches per epoch   & 4     \\ 
\bottomrule
\end{tabular}}

\end{table}

\section{Domain Randomization Details}
\label{sec:appendix_dr}

To improve sim-to-real transfer, we applied domain randomization to various simulation parameters during training. We followed previous work \cite{margolis2022walktheseways} for randomizing the robot's physical attributes. For LiDAR perception, we introduced specific randomizations: random masks were applied to 10\% of the point cloud, assigning these points small distance values uniformly sampled from [0, 0.3]. Additionally, 10\% noise was randomly added to the measured LiDAR distances.

The specific parameters and their randomization ranges are detailed in Table \ref{table:dr_params}. All attributes listed were uniformly sampled across all 4096 parallel environments during reinforcement learning.

\begin{table}[h!]
\centering
\caption{Parameters and Ranges for Domain Randomization}
\label{appendix:dr_params}
\label{table:dr_params}
\scalebox{0.8}{
\begin{tabular}{@{}l S[table-format=-1.3, table-number-alignment=center] 
                   S[table-format=1.3, table-number-alignment=center]@{}} 
\toprule
Parameter                          & \multicolumn{2}{c}{Range} \\
                                   & {Min} & {Max} \\ \midrule

LiDAR Point Masking Ratio          & \multicolumn{2}{c}{10\% (Values $\in [0, 0.3]$)} \\ 
LiDAR Distance Noise Ratio         & \multicolumn{2}{c}{10\%} \\ 
Added Mass (kg)                    & -1.0  & 5.0   \\
Payload Mass (kg)                  & -1.0  & 3.0   \\ 
Center of Mass x (m)               & -0.1  & 0.1   \\
Center of Mass y (m)               & -0.15 & 0.15  \\
Center of Mass z (m)               & -0.2  & 0.2   \\
Ground Friction Coefficient        & 0.40  & 1.00  \\ 
Ground Restitution                 & 0.00  & 1.00  \\ 
Motor Strength (Scale Factor)      & 0.8   & 1.2   \\
Joint Calibration Offset (rad)     & -0.02 & 0.02  \\ 
Gravity Offset (m/s$^2$)           & -1.0  & 1.0   \\ 
Proprioception Latency (s)         & 0.005 & 0.045 \\ \addlinespace 

\bottomrule
\end{tabular}
}
\end{table}

\section{Taichi Lidar Efficiency}
\label{appendix:Lidar_Efficiency}
We tested the performance of Taichi Version on three different computers, including a MacBook.Our program is cross-platform like MuJoCo,Genesis,Gazebo,Isaac sim/gym.

In scenes with fewer geoms (lower 200), simulating with 115,200 rays can achieve 500Hz+ simulation efficiency, which is really fast! Most of the time is spent in the preparation process, with a large proportion (more than60\%)
\begin{figure}[htbp!]
\centering
\setlength{\abovecaptionskip}{0cm}
\setlength{\belowcaptionskip}{0cm}
\includegraphics[width=1\linewidth]{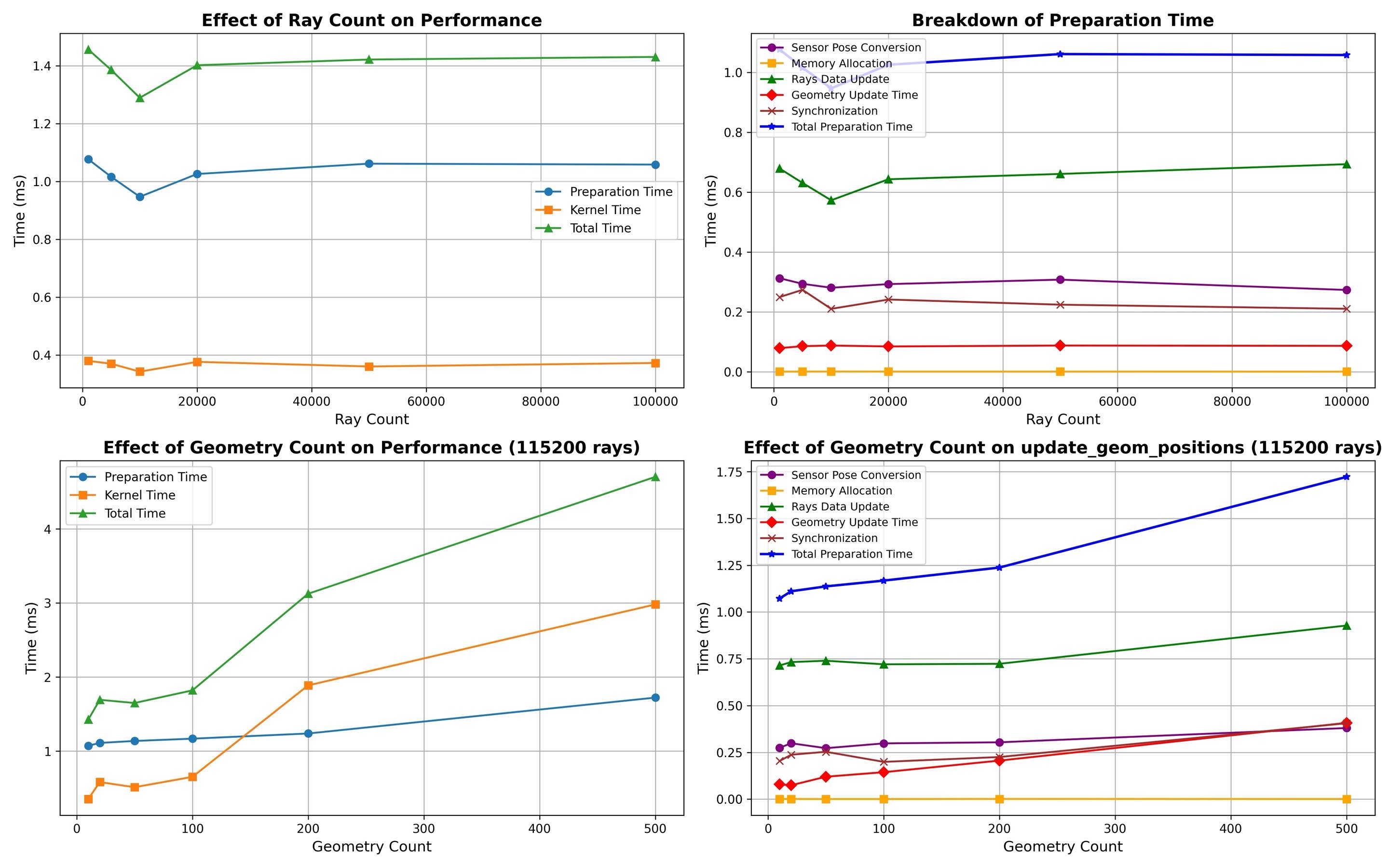}
\vspace{-0.2in}
\label{fig:Lidar_pattern}
\end{figure}

\end{document}